\documentclass[10 pt, conference]{IEEEtran}
\usepackage{graphics} 
\usepackage{amsmath} 

\usepackage{array, multirow}
\usepackage{graphicx}

\usepackage{tikz}
\usepackage{pgfplots}
\usetikzlibrary{backgrounds,arrows,calc,positioning,shapes, decorations.markings}

\usepackage{subcaption}
\usepackage{nohyperref}
\usepackage{url}
\usepackage{flushend}

\usepackage{soul}

\usepackage[normalem]{ulem}

\graphicspath{{resources/}}

\newcommand{\norm}[1]{\lVert #1 \rVert}

\title{\LARGE \bf
Depth-Based Visual Servoing Using Low-Accurate Arm}

\author{%
\IEEEauthorblockN{%
Ludovic Hofer\IEEEauthorrefmark{1},
Michio Tanaka\IEEEauthorrefmark{2},
Hakaru Tamukoh\IEEEauthorrefmark{2},
Amir Ali Forough Nassiraei\IEEEauthorrefmark{2},
Takashi Morie\IEEEauthorrefmark{2}}
\IEEEauthorblockA{\IEEEauthorrefmark{1}LaBRI, University of Bordeaux,
ludovic.hofer@labri.fr}
\IEEEauthorblockA{\IEEEauthorrefmark{2}Kyushu Institute of technology,
$\{$tamukoh,nassiraei,morie$\}$@brain.kyutech.ac.jp}}

\begin{document}

\maketitle
\thispagestyle{empty}
\pagestyle{empty}

\begin{abstract}
This paper proposes a visual-servoing method dedicated to grasping of daily-life
objects.
In order to obtain an affordable solution, we use a low-accurate robotic
arm. 
Our method corrects errors by using an RGB-D sensor.
It is based on SURF invariant features which allows us to perform object
recognition at a high frame rate. 
We define regions of interest based on depth
segmentation, and we use them to speed-up the recognition and to improve
reliability. 
The system has been tested on a real-world scenario. In spite of the lack of
accuracy of all the components and the uncontrolled environment, it grasps
objects successfully on more than 95\% of the trials.
\end{abstract}


\section{Introduction}

Visual servoing refers to robot control based on visual information. Computer
vision is used to analyze the visual data acquired by the robots. The data are
provided by one or several cameras, which can be placed directly on the robot
manipulator (eye-in-hand), on a mobile robot or somewhere fixed in the
workspace.
Grasping objects with a robot in a daily environment is a complex task that
service robots have to perform. The grasping problem refers to
choosing an optimal final state for the manipulator and the forces to apply
with its fingers.

The focus of this article is about the robustness of visual servoing based on
depth information toward a low-accurate robotic arm.  In particular, the choice
of the final state of the gripper is simplified by considering only cylindrical
objects and grasping them perpendicularly to the revolution axis of the
cylinder.  

In this paper, we present an entire grasping system.  In contrast to
the system used in~\cite{horaud98}, our goal is to grasp an object used in the
daily life such as a bottle. While our method can be extended to support other
situations, we make the following assumptions in our experiments: the object
is mostly cylindrical, its size is known and the background is not dark. The
system is designed to be embedded on service robots. Therefore, the sensor is
placed above the basis of the arm, on the same mobile platform. Since
affordability is a crucial point when designing service robots, the system has
been tested with a low-accurate robotic arm. While in article such
as~\cite{Bohg09} and~\cite{Kootstra12} simulation or industrial robots are used
to test the algorithms, these methods of evaluation do not fit the purpose of
this paper.

\begin{figure}
  \centering
  \includegraphics[width=.45\textwidth]{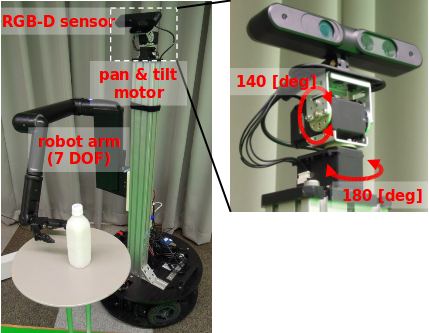}
  \caption{\label{fig:robot}The robot used for the experiments}
\end{figure}

The experiments of this system were performed using a 7 DOF (Degrees Of Freedom)
robotic arm, the {\em iARM} manufactured by {\em Exact Dynamics}. This robotic
arm is designed to be used by disabled people who control it using their own
vision to move it to a target position. Therefore, it was designed to be
controlled by a vision-based control loop and not only with a target
position. When controlled in position, it presents an accuracy of about 3 cm.
It is important to note that the error has a low repeatability, and therefore it
is not possible to learn a deterministic model for the error.
The RGB-D sensor used was an {\em Asus Xtion PRO LIVE} and its
orientation was controlled by servomotors from {\em Dynamixel}. 
Figure~\ref{fig:robot} shows the robot used and
the possible moves of the pan-tilt. In
this paper, the model of the robot is constructed based on the distances between
its different parts such as the arm basis, the camera and the servomotors.

Visual servoing using eye-in-hand cameras has been proven very
successful~\cite{Maxim12,Chaumette06,Hong95,Recatala04,Tsui11} even on moving
targets. However, most robotic arms come with no vision system, and the
integration of an eye-in-hand camera is problematic, especially if the arm has
to go through narrow spaces or if there is a collision risk. 
Moreover, cable of the camera
being disconnected was responsible for 13 failures out of 198 grasping trials
in~{\cite{Tsui11}}. In our approach, we use a camera installed on a pan-tilt mount in
order to avoid such problems. While we propose a tracking approach based on
identifying the fingers of the grippers, other alternatives based on depth
images exist such as~{\cite{Teuliere14}}, in which neither pose estimation nor 
feature extraction are required.

The main information used by our algorithm is noisy depth measurement results
provided by the RGB-D sensor. Our approach does not aim toward optimal control,
but toward real-time control. Actually, it has been proven in~\cite{Malis10} that
optimal control is difficult in presence of noise in the depth measurement.
Moreover, optimal control is suited for industrial environments while this paper
focuses on service robots. We use the color information to extract and match
SURF features~\cite{Bay08} while in~\cite{Maxim12}, SIFT
features~\cite{lowe99} are used. The architecture of a mobile manipulator
using SURF features to estimate the pose of objects is presented in~\cite{Song11}.
We use depth-based segmentation to obtain different regions of interest, thus
allowing us to run the local features matching on a subset. This allows us to reduce
the computation time and improve the accuracy of the position
estimation~\cite{Anh12}.

We measure the robustness of the system against errors in the model by
artificially adding errors in it. According to the results provided in~\cite{Malis10},
it shows that real-time control is an appropriate solution.

This paper makes four contributions:
\begin{enumerate}
\item An algorithm to detect regions of interest in depth images provided by
  RGB-D sensor.
\item A solution to keep a high frame rate while using a large set of SURF
  features~\cite{Bay08} for object recognition in real time.
\item A method of real-time tracking of the gripper based on regions of interest.
\item An experiment showing the accuracy improvement when visual servoing is
  used.
\end{enumerate}

The reliability of the proposed system has been confirmed by experimental
results, as well as its robustness against errors in the model. Therefore,
the designed system gives satisfying results, even when using low-accurate
robotic arms.

This paper is organized as follows: Section~\ref{sec:RelatedResearch} discusses
the related work in the areas of vision-based grasping and object recognition.
The detection method of regions of interest for object detection is proposed in
Section~\ref{sec:ROI}. The probabilistic database of references is presented in
Section~\ref{sec:DB}. In Section~\ref{sec:Filter}, we discuss the filter
proposed to
estimate the position of the object. Section~\ref{sec:Gripper} describes the
system proposed to track the position of the gripper in real-time.
The experimental scenario and an overview of the results are presented in
Section~\ref{sec:Experiments}. Finally, a conclusion is given in
Section~\ref{sec:Conclusion}.

\newcommand{\graphwidth}{.6\textwidth}
\newcommand{\graphheight}{4cm}
\newcommand{\graphfontsize}{\small}
\newcommand{\graphlegendsize}{\scriptsize}
\newcommand{\graphylabpos}{(0.08,0.5)}
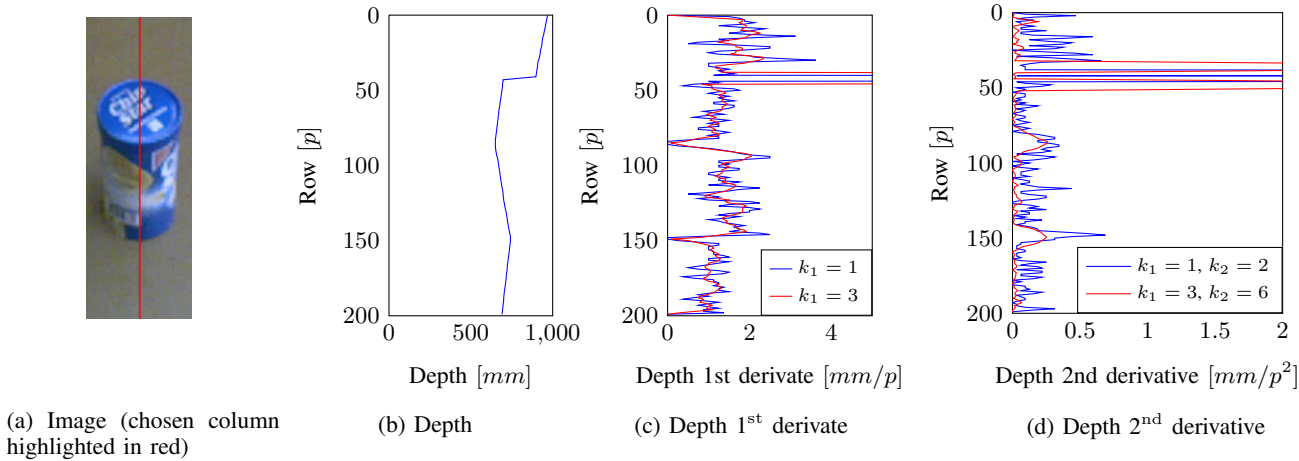
\begin{figure*}
  \centering
  \begin{subfigure}[t]{0.2\textwidth}
    \centering
    \vspace{-5.05cm}
    \includegraphics[height=\graphheight]{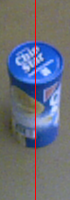}
    \vspace{1cm}
    \caption{\label{fig:CDSrc}Image (chosen column highlighted in red)}
  \end{subfigure}
  \begin{subfigure}[t]{0.2\textwidth}
    \begin{tikzpicture}
\begin{axis}
  [ 
    major x tick style=transparent,
    major y tick style=transparent,
    ylabel={Row $[p]$},
    y label style={at={\graphylabpos}},
    xlabel={Depth $[mm]$},
    no markers,
    ymin=0, ymax = 200,
    y dir = reverse,
    xmin=0, xmax =1000,
    xtick= {0,500,1000},
    width=\graphwidth,
    height=\graphheight,
    scale only axis,
    font=\graphfontsize
  ]
    \addplot table [col sep=semicolon,x index=1, y index=0]
             {resources/mixedDiff/DepthColumn.csv};
\end{axis}
\end{tikzpicture}
    \vspace{-0.42cm}
    \caption{\label{fig:CDDepth}Depth}
  \end{subfigure}
  \begin{subfigure}[t]{0.25\textwidth}
    \begin{tikzpicture}
\begin{axis}
  [ 
    major x tick style=transparent,
    major y tick style=transparent,
    ylabel={Row $[p]$},
    y label style={at={\graphylabpos}},
    xlabel={Depth 1st derivate $[mm/p]$},
    no markers,
    ymin=0, ymax = 200,
    y dir = reverse,
    xmin=0, xmax =5,
    legend image post style={xscale=0.5},
    legend entries={$k_1=1$, $k_1=3$},
    width=\graphwidth,
    height=\graphheight,
    scale only axis,
    legend style={at={(1.0,0.0)},anchor=south east,font=\graphlegendsize},
    font=\graphfontsize
  ]
    \addplot table [col sep=semicolon,x index=2, y index=0]
             {resources/mixedDiff/DepthColumnV2.csv};
    \addplot table [col sep=semicolon,x index=4, y index=0]
             {resources/mixedDiff/DepthColumnV2.csv};
\end{axis}
\end{tikzpicture}
    \vspace{-0.42cm}
    \caption{\label{fig:ColumnDer1}Depth 1$^{\mathrm{st}}$ derivate}
  \end{subfigure}
  \begin{subfigure}[t]{0.33\textwidth}
    \begin{tikzpicture}
\begin{axis}
  [ 
    major x tick style=transparent,
    major y tick style=transparent,
    ylabel={Row $[p]$},
    y label style={at={\graphylabpos}},
    xlabel={Depth 2nd derivative $[mm/p^2]$},
    no markers,
    xmin=0, xmax = 2,
    ymin=0, ymax = 200,
    y dir = reverse,
    legend entries={
      {$k_1=1$, $k_2=2$},
      {$k_1=3$, $k_2=6$},tmp},
    width=\graphwidth,
    height=\graphheight,
    scale only axis,
    legend style={at={(1.0,0.0)},anchor=south east,font=\graphlegendsize},
    font=\graphfontsize
  ]
    \addplot table [col sep=semicolon,x index=6, y index=0]
             {resources/mixedDiff/DepthColumnV2.csv};
    \addplot table [col sep=semicolon,x index=8, y index=0]
             {resources/mixedDiff/DepthColumnV2.csv};
\end{axis}
\end{tikzpicture}
    \caption{\label{fig:ColumnDer2}Depth 2$^{\mathrm{nd}}$ derivative}
  \end{subfigure}
  \caption{\label{fig:ColumnDerivatives}Example of derivatives estimations for a
    column of an image}
\end{figure*}
\section{\label{sec:RelatedResearch}Related research for object recognition}

Object recognition paradigm has radically evolved since the apparition of the
SIFT algorithm~\cite{lowe99}. While object recognition was previously based on
template matching, it is now essentially based on invariant features. However
extracting and matching SIFT features was time consuming. Several new types of
features have been proposed by the computer vision community. The SURF
algorithm~\cite{Bay08} outperforms SIFT in both quality and time consumption.
The use of binary descriptors has been brought by BRISK~\cite{Leutenegger11}.
Although it is less robust to changes of factors, it needs dramatically less
computational resources than SURF. A fast approximation method for the matching problem
has been developed as a part of the library presented
in~\cite{Muja14}. Finding a homography from the set of matches is usually done
by using RANSAC~\cite{Fischler81}. Reducing the area of the image to a region of interest
containing the object improves the quality of the results and decrease the
computation time~{\cite{Anh12}}.

A real-time plane segmentation using RGB-D sensors is detailed
in~\cite{Holz10}. Although it detects only plane objects, it presents an
interest for image segmentation since the computed information could be used
in addition to color information to improve color-based segmentation. A
scene interpreter based only on depth information is described
in~\cite{Rusu09}. It uses an accurate depth sensor and does not indicate the
computation time required but it shows a very high accuracy.

Different approaches to the grasping problem have been proposed in the robotics
community. In~\cite{Bohg09}, objects are divided into three categories: known
(model of the object is available), familiar (previous experience over similar
objects are available) and unknown (there is no knowledge about the object
available). The problem of using previous grasping experience to increase the
success rate is discussed in~\cite{Bohg09}. A method for the grasping of unknown
objects is presented in~\cite{Kootstra12}.  Reaching the grasping point by using
a visual based control loop is discussed in~\cite{horaud98}, this article uses
markers on the gripper to compute its orientation more easily. The arm we use
has an accuracy of about 3 cm. This is presumably much less accurate
than the robotic arms used in~\cite{Bohg09} and~\cite{Kootstra12}.

\section{\label{sec:ROI}Detecting regions of interest}

The system starts by detecting regions of interest (ROI) based on
the depth image captured by the RGB-D sensor. The advantage of using a depth-based
sensor over a stereo-camera is that it allows to detect objects independently of
their color difference from the background.

To perform efficient segmentation, two different kinds of edges need to be
detected.
\begin{itemize}
\item Sudden changes in depth, when the object hides a farther background.
  It corresponds to a local peak in the first derivative.
\item Changes of surface orientation, when there is no noticeable difference of
  depth between the object and the plane on which it is lying.
\end{itemize}
The differences between these two kinds of edges can be observed in
Fig.~\ref{fig:ColumnDerivatives}. Units used are $mm$ for millimeters and $p$
for pixel. The image used is shown in Fig.~\ref{fig:CDSrc}. The column of the
image which is used as an example is colored in red. The corresponding depth
acquired by the sensor is shown in Fig.~\ref{fig:CDDepth}.

Edges are detected both horizontally and vertically, using approximations of the
first and second derivative of the depth image. The
kernels used for derivation are $K_h$ and $K_v$ as defined in Eq.~\eqref{eq:K_h}
and Eq.~\eqref{eq:K_v}. The effects of the values $k_1$ and $k_2$ are shown in
Fig.~\ref{fig:ColumnDer1} and Fig.~\ref{fig:ColumnDer2}. The values $k_1$ and
$k_2$ are used as $k$ for the first and the second derivations respectively. Kernels of
an odd length and height are used in order to apply them with a centered anchor.
This helps to avoid shift in edge detection.

\begin{equation}
  \label{eq:K_h}
  K_h (k) = [-1, 0_{1 \times 2k - 1}, 1]
\end{equation}

\begin{equation}
  \label{eq:K_v}
  K_v (k) = [-1, 0_{1 \times 2k - 1}, 1]^T
\end{equation}

Using a threshold on the absolute value of the first derivative allows us to detect
sudden depth changes as can be seen in Fig.~\ref{fig:ColumnDer1} at row 40.
However, the same process does not work for the second derivative. In order
to detect the edge around row 150 in Fig.~\ref{fig:ColumnDer2}, the threshold
required would be low enough to detect a very thick edge around the sudden
depth changes as in row 40. A mask is used for detecting the second
derivative in order to avoid detecting edges which have already been detected by
the threshold on the first derivative. This mask is obtained by performing a
{\em dilatation} (a well-known morphological process) on the edges already
detected. In order to reduce the number of remaining edges, an {\em erosion}
process is performed once the threshold has been applied.

Using the depth image shown in Fig.~\ref{fig:SPSrc}, horizontal and
vertical edges are detected using the first and second derivative. By gathering
them, we obtain the image shown in
Fig.~\ref{fig:SPUndilatedEdges}. A dilatation is then performed in order to
ensure that the contours of the object are connected as shown in
Fig.~\ref{fig:SPEdges}.
A set of superpixels is obtained by applying a floodfill algorithm as shown
in Fig.~\ref{fig:SPUnfilteredSP} (Floodfill algorithm from the computer vision
library {\em OpenCV} is used). A first filter based on the size of the
superpixel is applied to the set of superpixels obtained by applying the
floodfill algorithm in order to remove the background and noise. Then,
by using the depth information and the position of the pixels,
the superpixels properties (i.e the height and the width of the superpixel)
are computed. The obtained properties for each superpixel are
compared with the expected properties. If it
does not match, then the superpixel is discarded. The remaining superpixels
(shown in Fig.~\ref{fig:SPFilteredSP}) are the ROI obtained from the depth
image. Those regions are highlighted in red in the corresponding color image as
shown in Fig.~\ref{fig:SPTaggedImg}.

\newcommand{\SPImg}[3]
{
  \begin{subfigure}[t]{0.22\textwidth}
    \includegraphics[width=\textwidth]{mixedDiff/21_#3.png}
    \caption{\label{fig:SP#1}#2}
    \vspace{0.2cm}
  \end{subfigure}
}

\begin{figure}
  \centering
  \SPImg{Src}{Depth image}{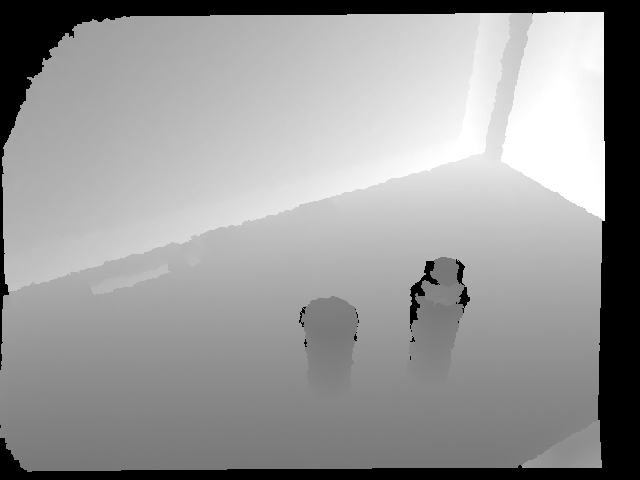}
  \SPImg{UndilatedEdges}{Detected contours}{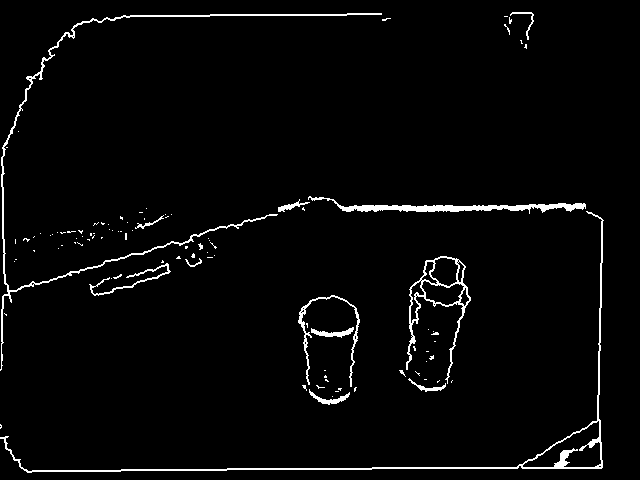}
  \SPImg{Edges}{Contours after dilatation}{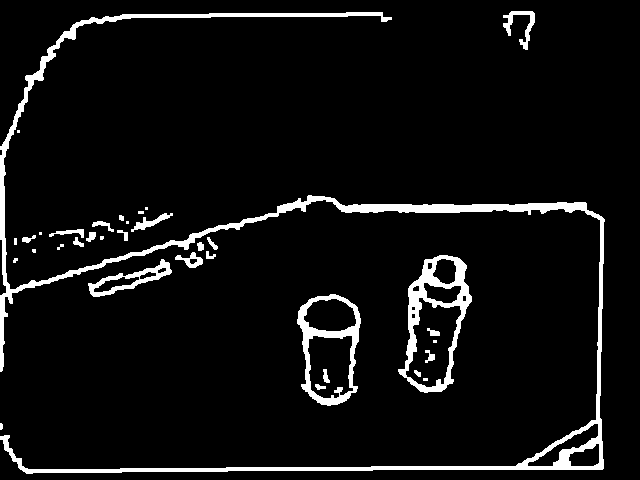}
  \SPImg{UnfilteredSP}{Superpixels}{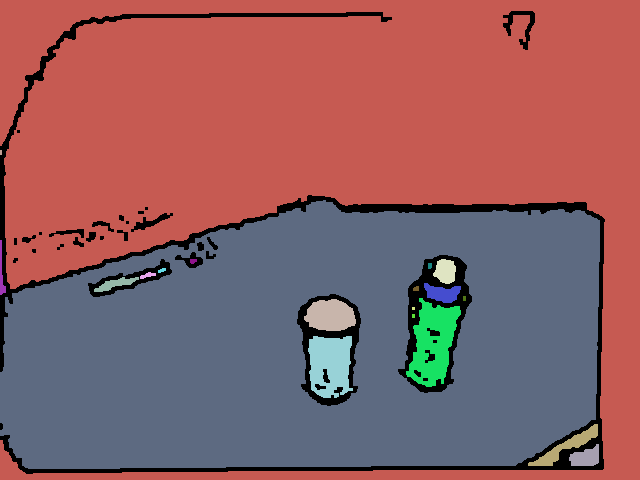}
  \SPImg{FilteredSP}{Filtered Superpixels}{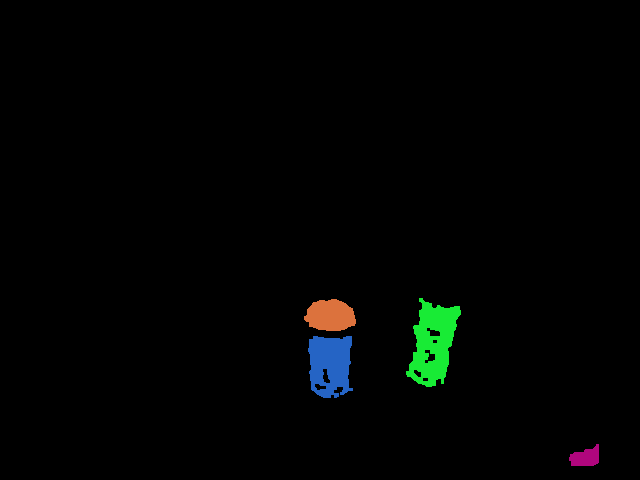}
  \SPImg{TaggedImg}{Detected ROI}{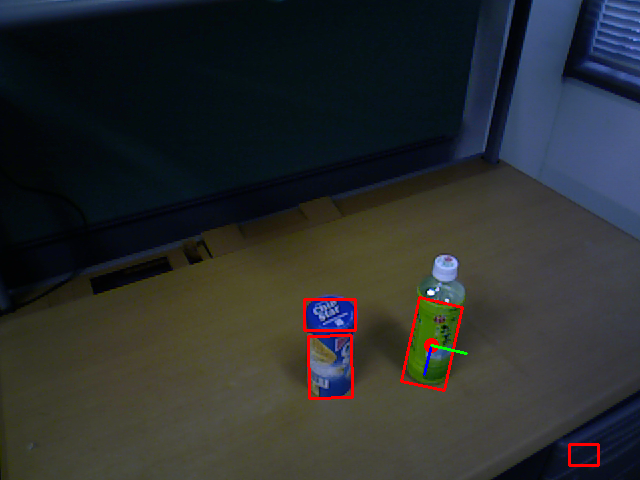}
  \caption{\label{fig:SP}Depth segmentation in six images}
\end{figure}


\section{\label{sec:DB}Probabilistic database of references}

In our system, each reference consists of a set of SURF features
computed in a ROI obtained through the process described above. A large
number of references are
used for every object in order to compensate for the small number of features
detected due to the size of the ROI. The database of references
stores all the references corresponding to an object. The main goal of making this
database is to allow the use of a large number of references while keeping a
low computation time in order to ensure real-time processing.

Once the set of ROI has been computed, each of the regions has to be compared with the
references by matching features of the references and those of the
input image. Using all the references
for each frame would lead to a low frame rate, which is inversely proportional to the
number of references in the database. In order to ensure that the frame rate does
not depend on the number of references, 
we only use a random subset of the references
with a bounded maximal size. In order to
increase progressively the chances to obtain a match, two rules are applied.
For each reference of the subset, if at least one of the ROI matches it,
then we increase the probability of selecting. On the other side, if none of
the ROI matches the reference, we decrease the probability of selecting it.

Let $n$ be the number of references of the database,
${R = \left ( r_1, \dots, r_n \right )}$ be a set of references of the database and
${W = \left ( w_{r_1}, \dots, w_{r_n} \right )}$ be a set of weights for the references.
Then, three properties are ensured by the database:
\begin{itemize}
\item The probability of obtaining reference $r$ is directly influenced by its
  weight $w_r$.
\item $\sum\limits_{w \in W} w = 1$.
\item Let $m$ and $M$ be the minimal and the maximal weights chosen by the user
  respectively. Then ${\forall w \in W, m \leq w \leq M}$. Ensuring this
  property avoids issues due to floating point approximation when the weight
  of a reference would be approximated by 1 or 0. It also allows to keep a
  non-zero probability of obtaining each reference.
\end{itemize}

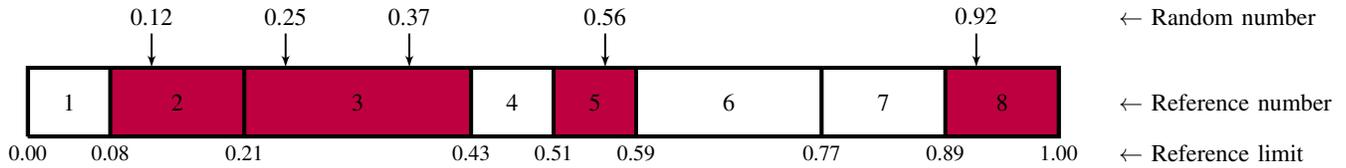
\begin{figure*}
  \vspace{0.05cm}
  \resizebox{\textwidth}{!}{\newcount\areaNo
\areaNo=1

\newcommand{\valArrow}[1]{
  \draw[->, thick] (#1,1.5) -> (#1,1);
  \draw (#1,1.5) node[above]{#1};
}

\newcommand{\drawArea}[3]{
 \draw[ultra thick, fill=#3] (#1,0) -- (#2,0) -- (#2,1) -- (#1,1) -- (#1,0);
 \draw (#2,0) node[below]{\small#2};
 \path (#1,0) edge[draw opacity=0] node {\the\areaNo} (#2,1);
 \advance\areaNo by 1
}

\begin{tikzpicture}[>=latex',xscale=15,yscale=1]
  \draw (0,0) node[below]{\small 0.00};
  \drawArea{0.00}{0.08}{white}
  \drawArea{0.08}{0.21}{purple}
  \drawArea{0.21}{0.43}{purple}
  \drawArea{0.43}{0.51}{white}
  \drawArea{0.51}{0.59}{purple}
  \drawArea{0.59}{0.77}{white}
  \drawArea{0.77}{0.89}{white}
  \drawArea{0.89}{1.00}{purple}

  \draw (1.05,0.0) node[below right]{$\leftarrow$ Reference limit};
  \draw (1.05,0.5) node[right]      {$\leftarrow$ Reference number};
  \draw (1.05,1.5) node[above right]{$\leftarrow$ Random number};

  \draw[ultra thick] (0,0) -- (1,0) -- (1,1) -- (0,1) -- (0,0);
  \valArrow{0.12}
  \valArrow{0.25}
  \valArrow{0.37}
  \valArrow{0.56}
  \valArrow{0.92}
\end{tikzpicture}}
  \caption{\label{fig:randomPicks}Picking randomly five elements out of the
    eight references from the database}
\end{figure*}

An example of picking references from the database is shown in
Fig.~\ref{fig:randomPicks}. Every reference is chosen at most once, even if two
random numbers point to the same reference. Therefore, when five references are
requested, a subset containing four references might be selected
(e.g. in Fig.~\ref{fig:randomPicks}). While
this might be considered as a drawback of this algorithm, it turns out to be
an advantage, because the average number of references in the random subset is
reduced when the
weight of a reference is significantly higher than $\frac{1}{n}$. Such a
situation happens only if this reference has been successfully matched with one
of the ROI. By reducing the number of references of the subset in this
situation, the required processing time highly decreases while the probability
for matching only slightly decreases.

The corrected weight for reference $z_r$ is computed by
using~\eqref{eq:weightSuccess} if matching has been successful
and~\eqref{eq:weightFailure} if it has not.
If the reference did not belong to the picked subset, then $z_r = w_r$. Two
different gains are defined, $G_s$ and $G_f$. When a match has been found,
$G_s$ is used. If no matches have been found, $G_f$ is used. Both gains have to
be positive and smaller than 1. These gains represent the learning rates.
Typical values used in experiments are $G_s = 0.3$ and
$G_f = 0.04$.

\begin{equation}
  \label{eq:weightSuccess}
  z_r = w_r + G_s (M - w_r)
\end{equation}

\begin{equation}
  \label{eq:weightFailure}
  z_r = w_r - G_f (w_r - m)
\end{equation}
After these updates, the sum of all $z_r$ might
be different from 1. We end up the update by normalization which depends on
${\delta = 1 - \sum\limits_{r \in R} z_r}$.

Let us define $u_r = M - w_r$ as the weight which can be added to a reference before
reaching the maximal value and $v_r = w_r - m$ as the weight which can be removed of a
reference before reaching the minimal value. Let us define $U = \sum\limits_{r \in R}u_r$
and $V = \sum\limits_{r \in R}v_r$ as the total weights which can be
added and removed from the references before reaching global saturation, respectively. Define $w'_r$ as the
final weight of a reference after normalization. If $\delta > 0$, then it
is computed according to Eq.~\eqref{eq:weightNorm1}, else it is computed
according to Eq.~\eqref{eq:weightNorm2}.

\begin{equation}
  \label{eq:weightNorm1}
  w'_r = z_r + \frac{u_r}{U} \delta
\end{equation}
\begin{equation}
  \label{eq:weightNorm2}
  w'_r = z_r + \frac{v_r}{V} \delta
\end{equation}

\section{\label{sec:Filter}Filtering the object position}

Even if the object to grasp is not supposed to move quickly, it is important to
use multiple detection results for determining the object position. This can
lower the false positive rate and improve the accuracy of the position estimation
at the same time. Since the object detection part has a non-zero probability of
confusing objects, it is necessary to remove the detected positions which do not
fit with the other results. This is achieved by remembering the last $l$ 
successful results of the object detection and removing the $j$ results which
bring the highest error, where $l$ and $j$ are two positive integers. 
The results of the object detection are provided
with an index of confidence based on the number of matched features, it provides
an estimate of the quality of the matching. In order to increase the impact of recent
information, a timestamp is associated with each result.

Each entry of the filter is composed of four different elements:
\begin{itemize}
\item computed position of the object; $h$
\item quality of the detection (index of confidence); $q$ 
\item associated timestamp; $t$
\item detected axis of the object; $a$
\end{itemize}
The state of filter $F$ is described formally by the following equation:
\begin{equation}
  \label{eq:filter}
  \begin{array}{lcl}
  F & = & \{e_1, \dots, e_l\}\\
    & = & \left \{ (h_1, q_1, t_1, a_1 ), \dots, (h_l, q_l, t_l, a_l) \right \}
  \end{array}
\end{equation}

Let $b_i$ be the weight of the entry $(h_i, q_i, t_i, a_i)$ at time $T$. Then,
the weight is calculated as follows:
\begin{equation}
  b_i =\frac{q_i}{T- t_i}
\end{equation}
Therefore, it 
is easy to see that the weight of an entry is proportional to its quality and
inversely proportional to the elapsed time since this entry appeared.

Let $E(F)$ be the object position estimated from the state of the filter,
and it is computed accordingly to Eq.~\eqref{eq:filteredValue}.
\begin{equation}
  \label{eq:filteredValue}
  E(F) =\frac{\sum\limits_{i=1}^l {h_i b_i}}{\sum\limits_{i=1}^l {b_i}}
\end{equation}

Let $Q(F)$ be the detection quality obtained from the state of the filter and
$\alpha$ be a constant used as a gain. Then, $Q(F)$ is computed
accordingly to Eq.~\eqref{eq:filteredQuality}.
\begin{equation}
  \label{eq:filteredQuality}
  Q(F) =\frac{\sum\limits_{i=1}^l \norm{E(F) - h_i} b_i}{\sum\limits_{i=1}^l {b_i}}
    \times \frac{\sum\limits_{i=1}^l \frac{1}{T - t_i}}{\alpha}
\end{equation}

In order to remove eventual false positives, we rank all the entries $e_i$
according to the value $Q(F \setminus e_i)$, 
where $F \setminus e_i$ represents $F$ without the $i$-th component $e_i$. 
Then, we build a new set $F' = \{e'_1, \dots, e'_{l-j}\}$ by
removing from $F$ the $j$ entries which led to the
lowest value. Thus, the final filter value and quality are represented by
$E(F')$ and $Q(F')$, respectively. 
It is important to note that $F'$ needs to be updated when a
new entry is added.

\section{\label{sec:Gripper}Tracking the gripper position}

Gripper tracking is necessary to correct the mechanical errors when using
low-accurate robotic arms. However, it is still possible to use the estimation
of the position according to the arm controller to improve the efficiency of
tracking. We chose to detect both fingers of the gripper separately.

\begin{figure}
  \centering
  \includegraphics[width=.45\textwidth]{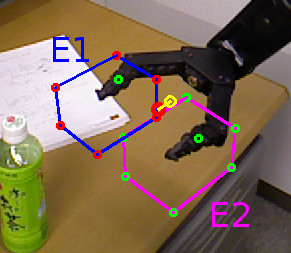}
  \caption{\label{fig:spaces}Projections of the possible arm finger positions on an image}
\end{figure}

First, we use the data provided by the arm controller to compute the bounding
boxes which are supposed to
contain the fingers. Then, we increase the size of both bounding boxes by taking
into account the maximal error of the arm $\epsilon$. The
obtained spaces $E_1$ and $E_2$ are described in Eq.~\eqref{eq:FingerPosWithMargin}.
The projection of those spaces on an image is shown in
Fig.~\ref{fig:spaces}.

\begin{equation}
  \label{eq:FingerPosWithMargin}
  \renewcommand{\arraystretch}{1.5}
  \begin{split}
    E_i = \left \lbrace
    {^G(x,y,z)} \left |
    \!\!
    \begin{array}{l}
      x \in [\frac{s_i g_w - f_w}{2} - \epsilon, \frac{s_i g_w + f_w}{2} + \epsilon]\\
      y \in [\frac{-f_t}{2} - \epsilon, \frac{f_t}{2} + \epsilon]\\
      z \in [-f_l - \epsilon, \epsilon]\\
    \end{array}
    \!\!\!\!
    \right .
    \right \rbrace
    \\
    \parbox[l]{.4\textwidth}{
      \vspace{0.5cm}
      Where:\\
      $^G(x,y,z)$ a point in the gripper basis\\
      $g_w$ the width of the gripper\\
      $f_w$ the width of a finger\\
      $f_t$ the thickness of a finger\\
      $f_l$ the length of the finger\\
      $\epsilon$ the maximal error accepted\\
      $s_i$ the side of the object: $s_1 = -1$ and $s_2 = 1$
    }
  \end{split}
\end{equation}

Once the spaces $E_1$ and $E_2$ have been computed, it is possible to use the basis
transformation to project them on the images from the RGB-D sensor.
The projection of $E_1$ and $E_2$ on the camera image is a polygon.
Let $x_m$ and $y_m$ be
the minimal $x$ and $y$ of the polygon respectively, and $x_M$ and
$y_M$ be the maximal $x$ and $y$ of the polygon respectively. Then
the regions of interest are defined by the rectangles which have the four following
corners: $(x_m, y_m)$, $(x_M, y_m)$, $(x_M, y_M)$ and $(x_m, y_M)$. Only these
regions of the image are selected and both the projections and the spaces $E_1$
and $E_2$ are used later in the gripper detection process.

The previous step allows us to compute a region of interest that should contain
the finger. Then two main steps are still requested to obtain the visual
position of the finger. Segmentation of the gripper needs to be performed and
the position of the finger has to be computed from the segmentation result.

In order to remove the pixels which do not belong to the space, we
check for every pixel if it belongs to the space. However, since this
requires to compute a basis transform for every pixel, two steps are performed
before it. First points which do not belong to the projection are removed as
shown in Fig.~\ref{fig:GripperProjection}. Then, color information are used to
remove pixels from the background as shown in Fig.~\ref{fig:GripperBlack}. In our
case, only black pixels are considered. Once all the conditions with a low
computational cost have been checked, the remaining pixels belonging
to the spaces $E_1$ and $E_2$ are checked. If it is not possible to
differentiate the background from the gripper by using color information, the
process is slowed because pixels belonging to space has to be checked for more
pixels.

A debugging image allowing to understand the space checking process is shown in
Fig.~\ref{fig:GripperBB}. Here, the {\em blue}, {\em green} and
{\em red} components of the image are respectively set to the maximal values if
the point matches the $^Gx$, $^Gy$ and $^Gz$ ranges defined
in Eq.~\eqref{eq:FingerPosWithMargin} (e.g. Pixels matching the  range $^Gx$
and $^Gz$ but not $^Gy$ are colored in purple which is the addition of blue and
red). Only the pixels colored in white are retained in the filtered image shown
in Fig.~\ref{fig:GripperFiltered}.

\newcommand{\gripperImgWidth}{.15\textwidth}
\newcommand{\gripperImgSpacing}{.1\textwidth}
\begin{figure}
  \vspace{0.17cm}
  \centering
  \begin{subfigure}[t]{\gripperImgWidth}
    \includegraphics[width=\textwidth]{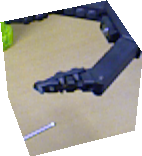}
    \caption{\label{fig:GripperProjection}Projection of the space on the image}
  \end{subfigure}
  \hspace{\gripperImgSpacing}
  \begin{subfigure}[t]{\gripperImgWidth}
    \includegraphics[width=\textwidth]{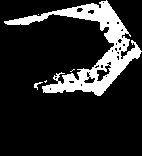}
    \caption{\label{fig:GripperBlack}Color matching pixels}
  \end{subfigure}
  \\[0.2cm]
  \begin{subfigure}[t]{\gripperImgWidth}
    \includegraphics[width=\textwidth]{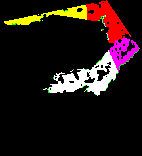}
    \caption{\label{fig:GripperBB}Space check}
  \end{subfigure}
  \hspace{\gripperImgSpacing}
  \begin{subfigure}[t]{\gripperImgWidth}
    \includegraphics[width=\textwidth]{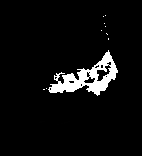}
    \caption{\label{fig:GripperFiltered}Filtered ROI}
  \end{subfigure}
  \caption{\label{fig:GripperFilteringROI}Filtering a region of interest}
\end{figure}

The image shown in Fig.~\ref{fig:GripperFiltered} is a binary one. This allows us
to run a floodfill algorithm to obtain a set of superpixels. Finally, only the
largest superpixel is retained, removing the noise which might have passed the
previous filtering steps.
If the obtained superpixel is large enough, the position of the finger is
calculated as the average of the remaining pixels. If there is no large enough
superpixel, the finger is considered as undetected.

A Kalman filter is used to estimate the final position of the gripper. It
uses a single position which is computed by using the results of the finger
detections. Since fingers can be undetected, it is necessary to handle different
cases.

\begin{itemize}
\item Both fingers detections fail: In this case, it is not possible to compute
  the gripper position from the image. A detection failure result is given to
  the Kalman filter.
\item One finger detection succeeds Let $\beta_t$ be the theoretical distance
  from the finger detected to the gripper. By assuming that the orientation
  error of the gripper is small, it is possible to estimate the visual position
  of the gripper $g_v$ by using the visual position of the finger $f_v$. The
  resulting value $g_v$ is given to the Kalman filter.
  \begin{equation}
    g_v \approx f_v + \beta_t
  \end{equation}
\item The two fingers are detected: The estimated position of the finger
  is the average of the positions of the two fingers. If the distance between
  the two fingers positions matches the expected distance, this value is
  given to the Kalman filter. If the difference between the visual distance of
  the two fingers and the theoretical distance is too large, a detection failure
  is given to the Kalman filter instead.
\end{itemize}

When a failure detection is sent to the Kalman filter, the values of the
associated covariance matrix are increased, increasing the uncertainty about
the gripper position and speed. While the uncertainty is below a given threshold,
the camera focus on the estimated position and a correction is applied to the
requested position using a simple proportional action. If the uncertainty is
too high, then the gripper stops and waits for the filter to stabilize.


\section{\label{sec:Experiments}Experimental results}

We conducted the experiments about the robustness of our grasping system by
running a set of trials. The database used for these experiments contained 50
RGB-D images for each of the five objects used and was built by moving objects
inside the grasping area. While several objects were presented in the database,
only one was used for the experiments. Closing the gripper to grasp the object
was achieved by requesting an finger space slightly lower than the diameter of
the object. All the trials were made on a laptop using a windows operating
system, the processor used was an Intel i3-2310M (2.1~GHz, Dual-core). We used
the C++ implementation of \emph{OpenCV 2.4}.

The task designed to evaluate the performance was as follows: the object was
placed alone on a table, the robotic arm was always starting at the same
position, gripper initially opened. When the grasping task was launched, the
reference database for visual detection was loaded, and the score of all the
references were set as equivalent in order to avoid influence from the previous
detections, the only input from the user side was the name of the object which
needs to be grasped. The evaluation task was to detect the object, grasp it and
lift it 15 cm above the original position. Neither the orientation of the object
nor its position with respect to the robot were fixed. However, the object was
placed inside a circle with a diameter of 20 cm.
We assumed six possible results for a grasping task:
four failures and two successes.

\begin{description}
\item[Object not detected:] \hfill \\
  After 30 s, if the object has not been detected by the vision module,
  then the object detection is considered as a failure and the trial is
  stopped.
\item[Gripper lost:] \hfill \\
  If the gripper detection system cannot manage to track the gripper, the
  system will not be able to bring the gripper to destination in less than
  30 s. In this case, the gripper is considered as lost and the trial
  is stopped.
\item[Grasping failed:] \hfill \\
  On the way to the destination, the gripper might collide with another
  object. The gripper might also close while it is not on the object. In both
  cases, the grasping part is considered as a failure.
\item[Lifting failed:] \hfill \\
  Sometimes, the gripper closes on the object but not
  accurately enough to lift it and the object falls while trying to lift it.
  In this case, the lifting is considered as a failure.
\item[Object touched:] \hfill \\
  If the object has been lifted with success but there was a
  contact between the gripper and the object before the gripper started to
  close, then the trial is considered as a partial success.
\item[Object not touched:] \hfill \\
  If the object has been lifted with success and there was no
  contact between the gripper and the object before closing the gripper, then
  the trial is considered as a success.
\end{description}
Distinction between failures allows to know which part of the system is the
weakest, and distinction between successes allows us to estimate the accuracy of
the system.

Two different approaches were compared:
\begin{description}
\item[Non-Visually Guided Grasping (NVGG):] \hfill \\
  Tracking of the
  gripper is not activated and the estimate of the gripper position is directly
  given by the controller of the robotic arm.
\item[Visually Guided Grasping (VGG):] \hfill \\
  This method uses the detection system
  described previously to estimate the gripper position.
\end{description}
Both use the object recognition system proposed in this article.

It is shown in TABLE~\ref{table:RbyMethod} that without injected error, both
methods succeeded to lift the object in more than 95\% of the trials. The main
advantage of using VGG is that the rate of success without touching the
object before closing the gripper is significantly higher.
The frequency of \emph{Object not touched} is 70\% when using VGG,
while it is only 54\% when using NVGG.

This difference of accuracy is highlighted with the results of another
experiment where errors were introduced in the distance between the arm and the
camera. The error vector which was added to the model had a length of 4
cm, and was computed randomly at the beginning of each trial. The
results are shown in TABLE~\ref{table:RwithError} and it illustrates that when the model
of the robot is not accurate, VGG strongly outperforms NVGG.
The frequency of \emph{Success} is 72\% when using VGG while it is
only 48\% when using NVGG. Moreover, the frequency of \emph{Object not touched} is
54\% when using VGG, while it is only 12\% when using NVGG.

\begin{figure}
  \vspace{0.15cm}
  \centering
  \includegraphics[width=.45\textwidth]{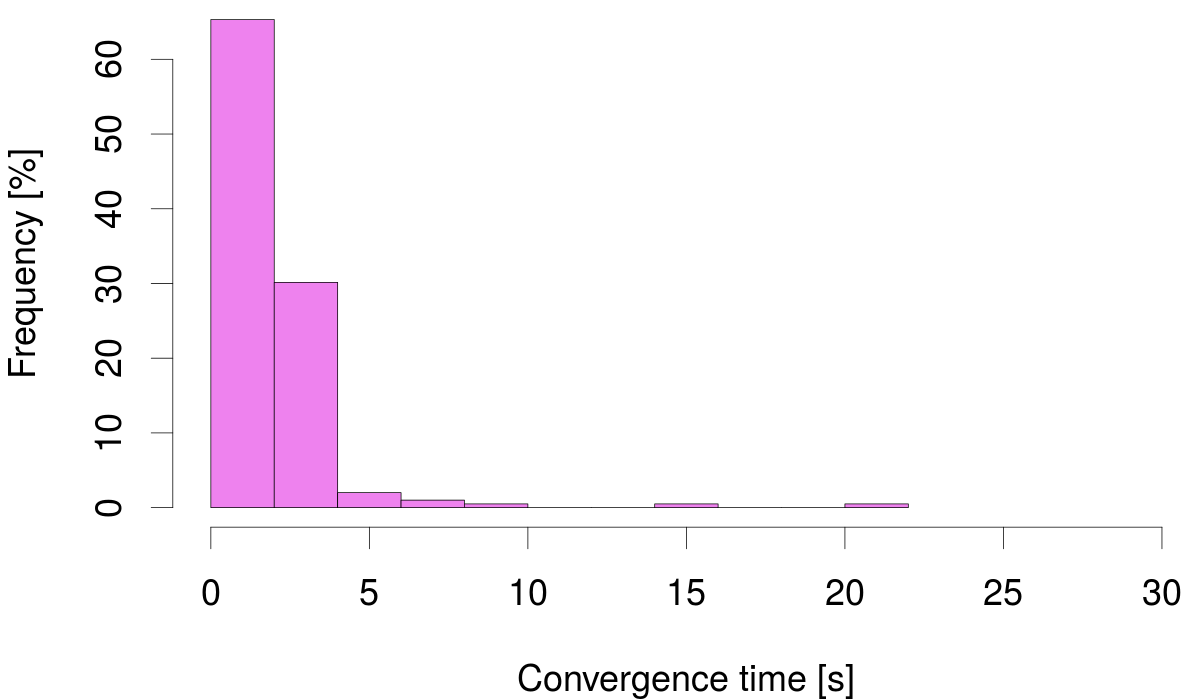}
  \caption{\label{fig:TCforDetection}Histogram of convergence time}
\end{figure}

The time needed to converge to an accurate estimate of the position of the object
was also recorded and is shown in Fig.~{\ref{fig:TCforDetection}}. This measure represents
the time required before the \emph{quality} of the filter grows above the
hand-tuned threshold, which ensures that the estimation of the object position
is reliable. In more than 60\%
of the cases, the time needed to detect the object and to estimate its position was
within 2 s. It was greater than 4 s in less than 10\% of the
cases. However, there are still isolated cases when the detection takes more than
20 s. This is due to the fact that the set of references did not cover all
the possible cases. Sometimes the matching between the current state of the
object and the references was hard to establish because there was not
sufficient references. 

The matching time for each frame using the probabilistic
database proposed here was compared to the required time per frame when using all
references in Fig.~{\ref{fig:MatchingTime}}. The proposed method allows to use
only a subset of references, ensuring that the matching time is not
proportional to the number of references. The data were computed from 20
different videos of 10 s and only the convergence time was examined.

\begin{table}
  \vspace{0.15cm}
  \caption{\label{table:RbyMethod}Grasping results by method}
  \centering
  \begin{tabular}{c|l|c|c}
	& \bf Result & \bf Count & \bf Frequency\\
\hline
	\multirow{4}*{\rotatebox{90}{\hspace{0.1cm} \bf NVGG}}%
	& \underline{Success} & \underline{50} & \underline{100\%}\\
	& \hspace{0.5cm}\em \footnotesize Object not touched & \em \footnotesize 21 & \em \footnotesize 42\%\\
	& \hspace{0.5cm}\em \footnotesize Object touched & \em \footnotesize 29 & \em \footnotesize 57\%\\
\hline
	\multirow{5}*{\rotatebox{90}{\bf VGG}}%
	& \underline{Success} & \underline{49} & \underline{98\%}\\
	& \hspace{0.5cm}\em \footnotesize Object not touched & \em \footnotesize 35 & \em \footnotesize 70\%\\
	& \hspace{0.5cm}\em \footnotesize Object touched & \em \footnotesize 14 & \em \footnotesize 28\%\\
	& \underline{Failure} & \underline{1} & \underline{2\%}\\
	& \hspace{0.5cm}\em \footnotesize Object not detected & \em \footnotesize 1 & \em \footnotesize 2\%\\
\end{tabular}

\end{table}

\begin{table}
  \caption{\label{table:RwithError}Grasping results by method with 40 mm error
    in the model}
  \centering
  \begin{tabular}{c|l|c|c}
	& \bf Result & \bf Count & \bf Frequency\\
\hline
	\multirow{7}*{\rotatebox{90}{\bf NVGG}}%
	& \underline{Success} & \underline{24} & \underline{48\%}\\
	& \hspace{0.5cm}\em \footnotesize Object not touched & \em \footnotesize 6 & \em \footnotesize 12\%\\
	& \hspace{0.5cm}\em \footnotesize Object touched & \em \footnotesize 18 & \em \footnotesize 36\%\\
	& \underline{Failure} & \underline{26} & \underline{52\%}\\
	& \hspace{0.5cm}\em \footnotesize Object not detected & \em \footnotesize 3 & \em \footnotesize 6\%\\
	& \hspace{0.5cm}\em \footnotesize Grasping failed & \em \footnotesize 11 & \em \footnotesize 22\%\\
	& \hspace{0.5cm}\em \footnotesize Lifting failed & \em \footnotesize 12 & \em \footnotesize 24\%\\
\hline
	\multirow{7}*{\rotatebox{90}{\bf VGG}}%
	& \underline{Success} & \underline{36} & \underline{72\%}\\
	& \hspace{0.5cm}\em \footnotesize Object not touched & \em \footnotesize 27 & \em \footnotesize 54\%\\
	& \hspace{0.5cm}\em \footnotesize Object touched & \em \footnotesize 9 & \em \footnotesize 18\%\\
	& \underline{Failure} & \underline{14} & \underline{28\%}\\
	& \hspace{0.5cm}\em \footnotesize Gripper lost & \em \footnotesize 7 & \em \footnotesize 14\%\\
	& \hspace{0.5cm}\em \footnotesize Grasping failed & \em \footnotesize 3 & \em \footnotesize 6\%\\
	& \hspace{0.5cm}\em \footnotesize Lifting failed & \em \footnotesize 4 & \em \footnotesize 8\%\\
\end{tabular}

\end{table}

\renewcommand{\graphwidth}{.37\textwidth}

\begin{figure}
  \centering
  \begin{tikzpicture}
\begin{axis}
  [ 
    ylabel={Matching time $[ms]$},
    xlabel={Size of the database},
    legend entries={All references, Probabilistic database},
    legend style={at={(0.0,1.0)},anchor=north west,font=\graphlegendsize},
    no markers,
    ymin=0, ymax = 120,
    xmin=0, xmax = 100,
    xtick= {0,50,100},
    width=\graphwidth,
    height=5cm,
    scale only axis
  ]
    \addplot table [col sep=semicolon,x index=0, y index=1]
             {resources/MatchingTimes.csv};
    \addplot table [col sep=semicolon,x index=0, y index=2]
             {resources/MatchingTimes.csv};
\end{axis}
\end{tikzpicture}
  \caption{\label{fig:MatchingTime}Matching time by method}
\end{figure}
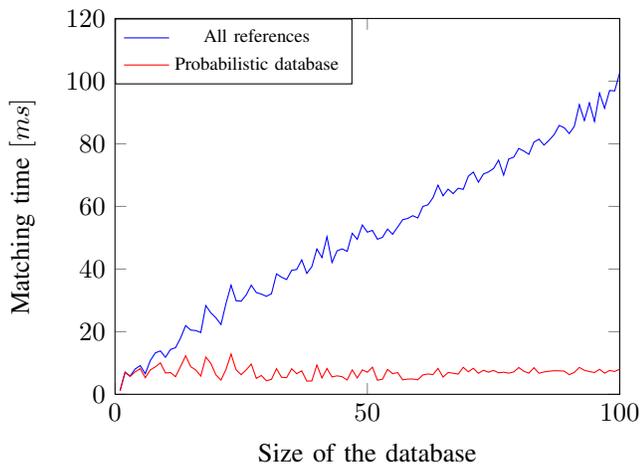

\section{\label{sec:Conclusion}Conclusion}

We presented an entire grasping system which shows a high success rate even
when the position of the gripper given by the model is not accurate. In contrast
to other approaches with high performances, our system offers a possibility of
using very low-accurate robotic arms. Since the cost of service robots is a crucial
issue, we believe that the proposed system is a major step toward their
globalization.

Amongst the further research into this system, we aim to improve the database of
references in order to develop the ability of learning new objects autonomously.
Another step which needs to be taken in order to provide a grasping system
adapted to daily life environment is collision avoidance when planning the
gripper trajectory.

\section*{Acknowledgments}

This work was done during an international internship supported by the 
Joint graduate school - Intelligent car and robotics course in Kitakyushu, Japan. 
The authors 
would like to thank Olivier Ly and Hugo Gimbert from the ``Laboratoire Bordelais
de recherche en Informatique'' (LaBRI) for their useful comments and their
support.

\vspace{0.55cm}

\bibliography{IEEEabrv,biblio/bibliography}

\end{document}